%% file: egpaper_for_review.tex
\documentclass[10pt,twocolumn,letterpaper]{article}

\usepackage{cvpr}
\usepackage{times}
\usepackage{epsfig}
\usepackage{graphicx}
\usepackage{amsmath}
\usepackage{amssymb}
\usepackage{subfiles}
\usepackage{titling}

\usepackage[breaklinks=true,bookmarks=false]{hyperref}

\cvprfinalcopy % *** Uncomment this line for the final submission

\def\cvprPaperID{5878} % *** Enter the CVPR Paper ID here

\date{}
\begin{document}

%%%%%%%%% TITLE
\title{Representation Similarity Analysis \\ for Efficient Task taxonomy \& Transfer Learning}

\author{Kshitij Dwivedi \; \; \; \; \; \;  \; \; \; \; Gemma Roig\\
Singapore University of Technology and Design\\
%8 Somapah Rd, Singapore 487372\\
{\tt\small kshitij\_dwivedi@mymail.sutd.edu.sg, 
gemma\_roig@sutd.edu.sg}
}
\maketitle
\begin{abstract}
Transfer learning is widely used in deep neural network models when there are few labeled examples available. The common approach is to take a pre-trained network in a similar task and finetune the model parameters. This is usually done blindly without a pre-selection from a set of pre-trained models, or by finetuning a set of models trained on different tasks and selecting the best performing one by cross-validation. 
We address this problem by proposing an approach to assess the relationship between visual tasks and their task-specific models.
Our method uses Representation Similarity Analysis (RSA), which is commonly used to find a correlation between neuronal responses from brain data and models. With RSA we obtain a similarity score among tasks by computing correlations between models trained on different tasks. Our method is efficient as it requires only pre-trained models, and a few images with no further training. We demonstrate the effectiveness and efficiency of our method for generating task taxonomy on Taskonomy dataset. 
We next evaluate the relationship of RSA with the transfer learning performance on Taskonomy tasks and a new task: Pascal VOC semantic segmentation. Our results reveal that models trained on tasks with higher similarity score show higher transfer learning performance. Surprisingly, the best transfer learning result for Pascal VOC semantic segmentation is not obtained from the pre-trained model on semantic segmentation, probably due to the domain differences, and our method successfully selects the high performing models.   
\end{abstract}

%%%%%%%%% BODY TEXT
\section{Introduction}
\begin{figure}[t]
\begin{center}
   \includegraphics[width=1\linewidth]{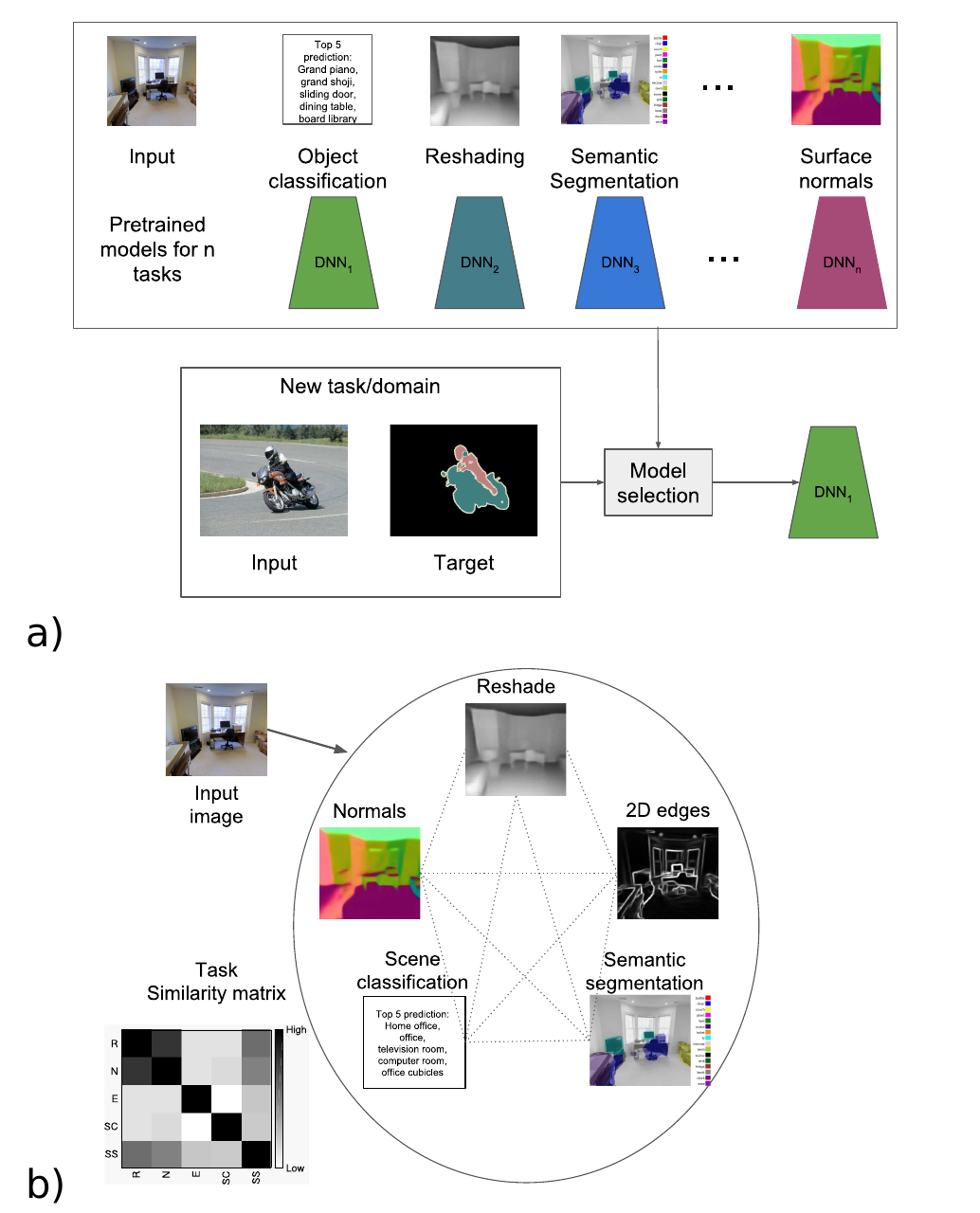}
\end{center}
   \caption{\textbf{Aims of this paper:} \textbf{a)} Deploy a strategy for model selection in transfer learning by \textbf{b)} Finding relationship between visual tasks. \vspace*{-3mm}  }
\label{fig1}
\end{figure}

%%Broader picture
For an artificial agent to perform multiple tasks and learn in a life-long manner, it should be able to re-utilize information acquired in previously learned tasks and transfer it to learn new tasks from a few examples. A solution to the aforementioned setting is to use transfer learning. Transfer learning allows to leverage representations learned from one task to facilitate learning of other tasks, even when labeled data is expensive or difficult to obtain.~\cite{ren2015faster,chen2017rethinking,long2015fully,girshick14CVPR}.

%% Intro to transfer learning
With the recent success of deep neural networks (DNN), these have become the \emph{ipso facto} models for almost all visual tasks~\cite{krizhevsky2012imagenet,simonyan2014very,he2016deep,zhao2017pspnet,he2017mask,zamir2018taskonomy}. The deployment of DNN has become possible mostly due to a large amount of available labeled data, as well as advances in computing resources~\cite{krizhevsky2012imagenet,simonyan2014very,he2016deep}. The need for data is a limitation that researchers have overcome by introducing transfer learning techniques. Transfer learning in DNN commonly consists of taking a pre-trained model in a similar task or domain, and finetune the parameters to the new task. For instance,~\cite{ren2015faster,girshick14CVPR} used a pre-trained model on ImageNet and finetuned it for object detection on Pascal VOC.  

%%Introduction to the problem setting
With a large number of pre-trained models (Figure~\ref{fig1}a) available, trained on a variety of vision tasks, it is not trivial how to select a pre-trained representation suitable for transfer learning.
To devise a model selection strategy, it is crucial to understand the underlying structure and relationship between tasks (Figure~\ref{fig1}b). If the relationship between different tasks is known, the model selection can be performed by evaluating similarity rankings of different tasks with a new task, using available pre-trained models. 

%%Taskonomy introduction
In a recent work, \cite{zamir2018taskonomy} modeled the relationship between tasks with a fully computational approach. They also introduce a dataset called Taskonomy, which contains labels of different visual tasks, ranging from object classification to edge occlusions detection. In this paper, we use the term Taskonomy for both the approach and the dataset from~\cite{zamir2018taskonomy}. 

%%Taskonomy details
 Taskonomy approach successfully computes the relationship between tasks. Yet, the relationship between a new task with an existing set of tasks is calculated with the transfer learning performance, which is tedious and computationally expensive. The performance on the new task is referred to transfer learning performance. To obtain the relationship of all previous tasks with the new task, Taskonomy approach also needs to compute the transfer learning performance on all the previous tasks using a model trained on the new task as a source. This defeats the purpose of not training a model from scratch for the new task, and all the procedure is computationally demanding as it is repeated for all the existing set of specific-task models. %In addition, if a task is in a new domain but the task exists in the set of pre-trained task-specific models, the Taskonomy approach assumes that the model trained on the same task will be the best choice for transfer learning. As we will show in our experiments, the same task may not be the best representation for transfer performance when it is a new domain. 
In this work, we address the above limitations by providing an alternative method to find the relationship between tasks.

%%Introduction to our approach
We propose a novel approach to obtain task relationships using representation similarity analysis (RSA). In computational neuroscience, RSA is widely used as a tool to compare brain responses with computational and behavioral models. Motivated by the success of RSA in neuroscience~\cite{kriegeskorte2008representational,Cichy2016,Khaligh-Razavi2014,Bonner,cichy2014resolving,MartinCichy2017,groen2018distinct}, we investigate the application of RSA in obtaining task similarities (Figure~\ref{fig1}b) and in transfer learning (Figure~\ref{fig1}a). Our approach relies on the assumption that the representations of the models that perform a related task will be more similar as compared to tasks that are not related, which we validate in our analysis. 

%% Advantages of our approach over taskonomy
In our approach, we compute the similarity scores using pre-trained task-specific models and a few examples. Thus, our RSA method only requires the representations of a few randomly selected images for all the tasks to compute the similarity, and we do not need to obtain transfer learning performance by finetuning on previous tasks' models. Further, we show in our results on Taskonomy dataset that task ranking similarity is independent of model size. Using small models trained with few samples for the existing tasks show similar results as the high performing models trained with all images. This allows to save computational time and memory, as well as it is more scalable to new tasks compared to Taskonomy approach.

%This indicates a direct relationship between similarity score and transfer learning performance which might differ depending on the domain, and proves the effectiveness of our method.

%%Results with advantages
We first validate the transfer learning applicability of our method on Taskonomy dataset. We find that for 16 out 17 Taskonomy tasks, the best model selected using RSA is in top-5 according to transfer learning performance. We also report results on Pascal VOC semantic segmentation task by analyzing the relationship of RSA similarity scores and the transfer learning performance. Our results show a strong relationship between RSA similarity score and transfer learning performance. We note that semantic segmentation model from Taskonomy dataset showed a lower similarity score than most of the 3D and semantic tasks, and a similar trend was observed in transfer learning performance. Our results suggest that in domain-shift, a model trained on the same task may not be the best option for transfer learning, and using our similarity score one can find a better model to achieve better performance. Using our RSA similarity scores method, we can select models with better transfer learning performance.

% we can delete the following. Intro is a bit too long and it is clear from the text and the figure.
%In summary, we make the following contributions: (i) We propose a novel approach to obtain a relationship between visual tasks, (ii) We show that our approach is model and task/domain agnostic, (iii) We show that task similarity score obtained using our approach is related to transfer learning performance.

\section{Related Works}
Here, we discuss the works that are most closely related to the aim of this paper, namely transfer learning in DNNs and Taskonomy.  Then, we briefly introduce the computational neuroscience literature that motivated our work. %Finally we discuss the areas where our proposed approach has potential applications.

\subsection{Transfer Learning}
The usual transfer learning approach in deep neural networks (DNNs) is to take a model pre-trained on a large dataset with annotations as an initialization of a part of the model. Then, some or all of the parameters are finetuned with backpropagation for a new task. The finetuning is performed because for most of the tasks there are insufficient annotations to train a DNN from scratch, which would lead to overfitting. Most of the works in the literature generally initialize the model parameters from a model pre-trained on  Imagenet~\cite{deng2009imagenet} dataset for image classification~\cite{krizhevsky2012imagenet,simonyan2014very,he2016deep,simonyan2014two,liu2015deep}. For example,~\cite{ren2015faster} use Imagenet initialized models for object detection on Pascal VOC,~\cite{long2015fully} use Imagenet initialized models for semantic segmentation. 

It has been noted in multiple works~\cite{mallya2018piggyback,sutskever2013importance,monfortmoments}, that the initialization plays a significant role in performance in transfer learning. Hence, a strategy is required to select models for initialization. Our proposed similarity-based ranking approach offers a solution to this problem, and as we discuss in the rest of the paper, tackles the limitations from  Taskonomy~\cite{zamir2018taskonomy}, which is one of the first attempts to tackle the model selection for transfer learning in DNN.   

\subsection{Taskonomy}
Our work is most closely related to Taskonomy~\cite{zamir2018taskonomy}, where the aim is to find the underlying task structure by computing the transfer performance among tasks. To achieve this goal, they create a dataset of indoor scene images with annotations available for $26$ vision tasks. The task set, which they refer as task dictionary, covers common 2D, 3D, and semantics computer vision tasks. Then, task-specific independent models are trained in a fully supervised manner for each task in the task dictionary.  They obtain a task similarity score by comparing the transfer learning performance from each of the task-specific models and computing an affinity matrix using a function of transfer learning performance. In this paper, instead of transfer learning performance, we rely on the similarity of the feature maps of the pre-trained models. Thus, we avoid additional training on pre-trained models to obtain transfer learning performance, saving computational time and memory, and still obtaining a meaningful relation with transfer learning performance as we will see in the results section.
%As we will see later, our method also prevents naively selecting a pre-trained model of the same task to which we want to transfer, as Taskonomy approach would do if another model from the dictionary is a better fit due to the domain dissimilarities.

\subsection{Similarity of computational models and brain responses}
In computational neuroscience, representation similarity analysis (RSA) is widely used to compare a computational or behavioral model with the brain responses. In~\cite{kriegeskorte2008representational}, RSA is used to compute similarities between brain responses in different regions of visual cortex with categorical models and computational vision models. In~\cite{Khaligh-Razavi2014}, the authors use several unsupervised and supervised vision models to show that supervised models explain IT cortical area better than unsupervised models, and \cite{MartinCichy2017} uses RSA to correlate the dynamics of the visual system with deep neural networks. 
We note that as the approach can be used to assess the similarity between a computational model and brain data, the approach can also be utilized to assess similarities between two computational models. RSA has been rarely used in the pure computational domain. Only in~\cite{mcclure2016representational} the RSA was introduced as a loss function for knowledge distillation~\cite{hinton2015distilling}, and in~\cite{MehrerCCN}, the consistency of RSA correlations with different random initialization seeds within the same model trained on CIFAR-10 \cite{krizhevsky2009learning} dataset is explored. However, RSA is still unexplored in comparing DNNs for assessing similarity among them.
%and did not validate on models trained on large scale datasets like Imagenet and Taskonomy. 
Our work introduces, for the first time, the use of RSA as a similarity measure to find the relationship between tasks,
%by using models trained on specific tasks,
and we believe it opens a new research line for the deep learning and computer vision.  

We use RSA similarity measure for two applications namely task taxonomy and transfer learning. Our approach is not limited to only these two applications and can be further applied in other computer vision problems. For instance, in multi-task learning~\cite{kokkinos2017ubernet,he2017mask,dharmasiri2017joint,li2015depth,dvornik2017blitznet} RSA could be used for deciding different branching out locations for different tasks, depending on their similarity with the representations at different depth of the shared root.

\begin{figure}[t]
\begin{center}
   \includegraphics[width=1\linewidth]{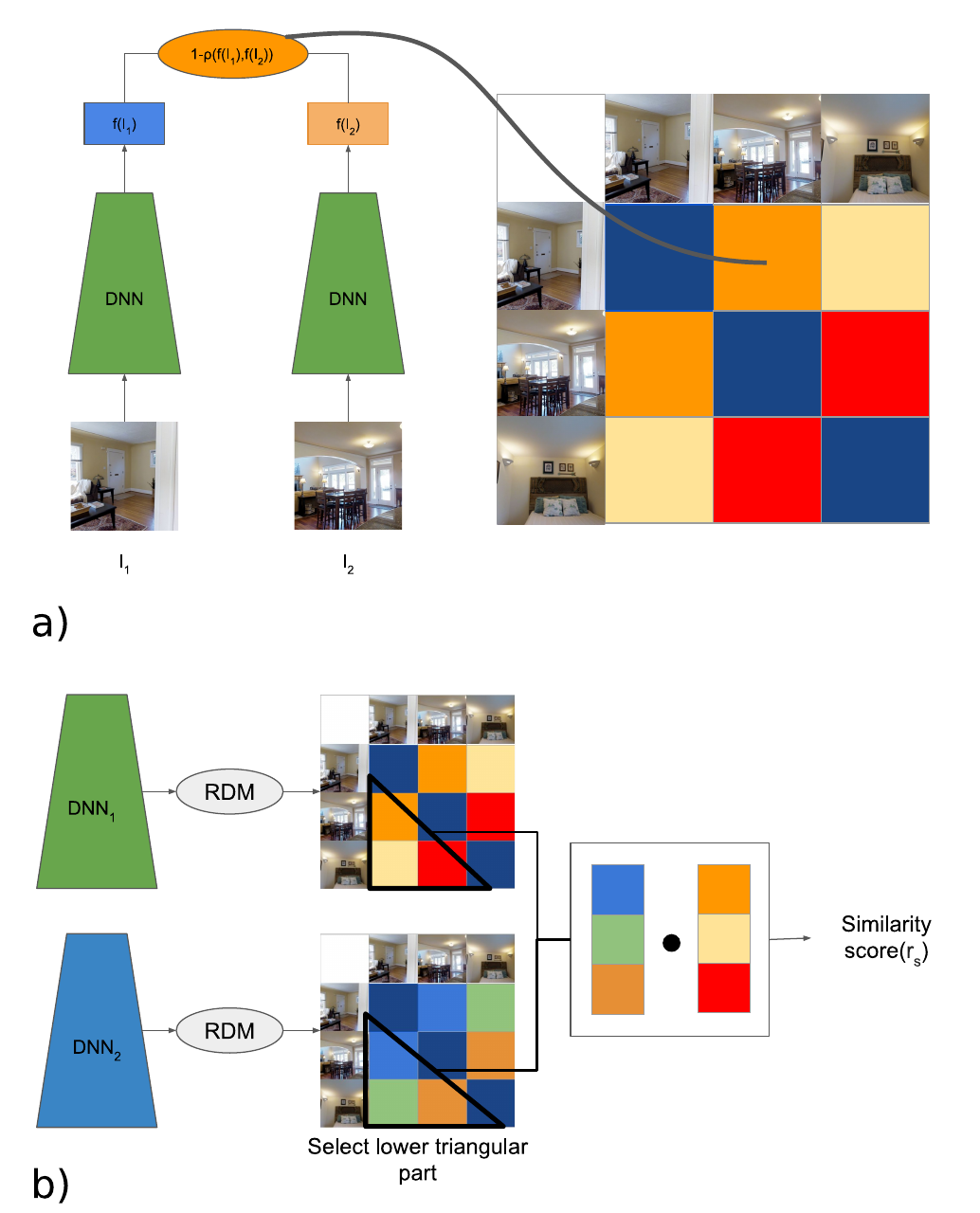}
\end{center}
   \caption{\textbf{Representation Similarity Analysis (RSA):} \textbf{a)} Representation dissimilarity matrices (RDMs) are generated by computing the pairwise dissimilarity (1 - Pearson's correlation) of each image pair in a subset of selected images. \textbf{b)} Similarity score: Spearman's correlation ($r_{s}$) (denoted with $\bullet$) of the low triangular RDMs of the two models is used as the similarity score. Here DNN\textsubscript{1} and DNN\textsubscript{2} refer to the models trained on task 1 and 2 respectively.\vspace*{-3mm}}
\label{fig2}
\end{figure}

%%%%%%
%RSA
%%%%%%

\section{Representation Similarity Analysis (RSA)}
\label{sec3}
Representation Similarity Analysis (RSA)~\cite{kriegeskorte2008representational}, illustrated in Figure~\ref{fig2}, is a widely used data-analytical framework in the field of computational neuroscience to quantitatively relate the brain activity measurement with computational and behavioral models.
In RSA, a computational model and brain activity measurements are related by comparing representation-activity dissimilarity matrices. The dissimilarity matrices are obtained by comparing the pairwise dissimilarity of activity/representation associated with each pair of conditions. 

In this work, we introduce RSA as a tool to quantify the relationship between DNNs and its application in transfer learning for model selection. We explain the steps to obtain the dissimilarity matrix for a computational model such as DNN in the following paragraph. 

\paragraph{Representation Dissimilarity Matrix (RDM)} We first select a subset of images as conditions for dissimilarity computation. For a given DNN, we then obtain the representation of each image by performing a forward pass through the model. For each pair of conditions (images), we compute a dissimilarity score $1-\rho$, where $\rho$ is the Pearson's correlation coefficient. The RDM  for this subset of conditions is then populated by the dissimilarity scores for each pair of conditions, see Figure \ref{fig2}a.

In our method, the RDMs computed for DNNs are used for obtaining the similarity between two computer vision tasks. Note that by using RDMs, the representation for different tasks can be of different length. The similarity is computed with the Spearman's correlation ($r_{s}$) between the upper or lower triangular part of the RDMs of the two DNNs. This is: 
$r_{s} = 1- {\frac {6 \sum d_i^2}{n(n^2 - 1)}}$, 
where $d_i$ is the difference between the ranks of $i\textsubscript{th}$ elements of the lower triangular part of the two RDMs in Figure~\ref{fig2}b, and $n$ are the number of elements in the lower triangular part of the RDM. 

The Spearman's correlation provides a quantitative measure of similarity between the task the DNNs were optimized for (Figure \ref{fig2}b). We explore the application of this similarity score in obtaining the relationship between computer vision tasks~\cite{zamir2018taskonomy}, and in transfer learning.

%%%%%%%%%%%%%%%%%%%%%%%%%%
%RSA for our applications
%%%%%%%%%%%%%%%%%%%%%%%%%%

\section{RSA for Task Taxonomy and \\Transfer Learning}
In this section, we introduce our RSA approach for getting a task taxonomy of computer vision tasks, as well as its application in transfer learning. We show the effectiveness of RSA for obtaining task similarity by answering three questions: 1) we investigate if we can group tasks into meaningful clusters based on task type using RSA on pre-trained task-specific models; 2) we analyze if the performance is important for computing task similarity or we can use a smaller subset of data with smaller suboptimal models; and 3) we investigate if the similarity we obtain using RSA is related to transfer learning.

\begin{figure*}[t]
\begin{center}
    \includegraphics[width=1\linewidth]{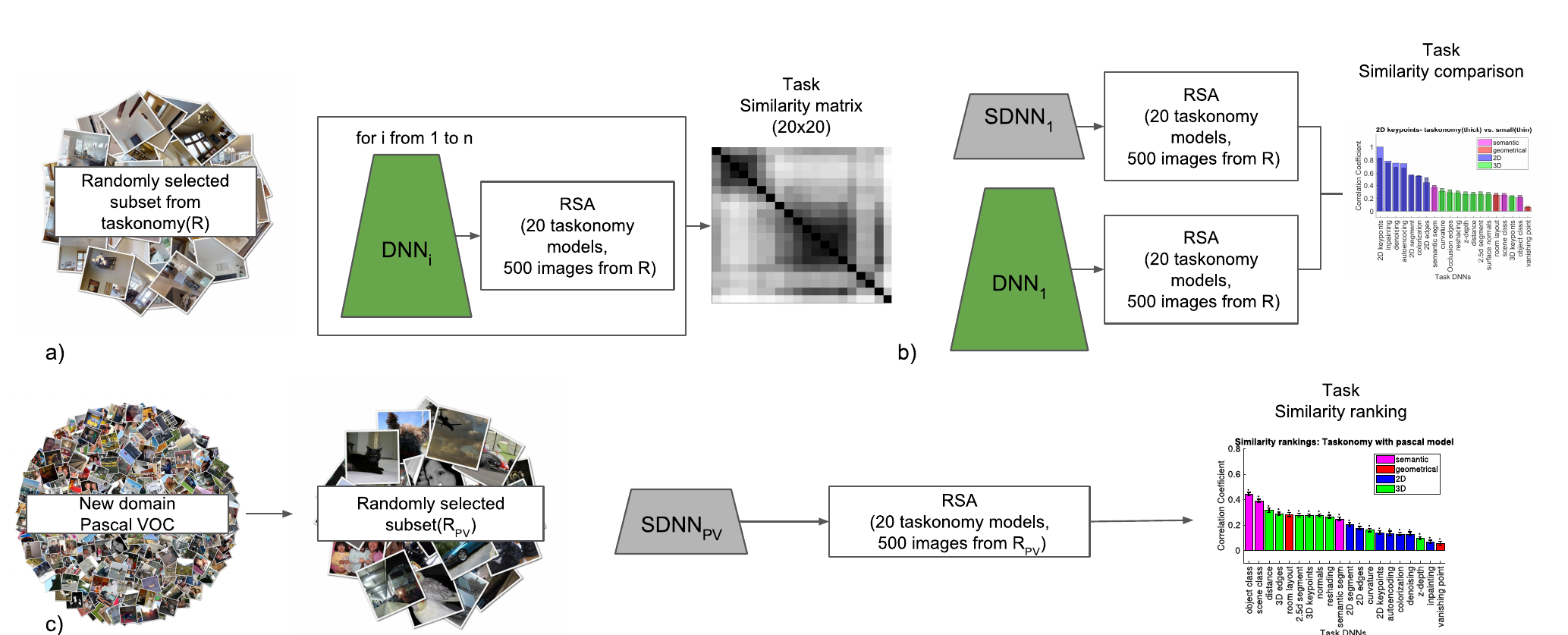}

\end{center}
\caption{\textbf{Our approach:} \textbf{a)} RSA of task-specific pre-trained DNN models (from Taskonomy) to compute a task similarity matrix, \textbf{b)} RSA of small model (SDNN) trained on small datasets and comparison with Taskonomy pre-trained models. \textbf{c)} RSA of small model (SDNN\textsubscript{PV}) trained on new task (Pascal VOC semantic segmentation) with Taskonomy pretrained models. \vspace*{-3mm}}
\label{fig3}
\end{figure*}

\subsection{Is task similarity related to task type?}
We validate our hypothesis that tasks similar according to RSA are grouped into clusters according to task type, for instance, 2D, 3D, semantic. To do so, we randomly select $500$ images from the Taskonomy dataset, and select $20$\footnote{we exclude Jigsaw task as it is unrelated to all other tasks} tasks from the task dictionary. Then, we compute the RDMs of the pre-trained models for each of the $20$ tasks using the task-specific representations of the $500$ sampled images, as described in section~\ref{sec3}. The task-specific representations are obtained by doing a forward pass on the pre-trained task-specific DNN models. With the resulting RDMs per task, we compute a pairwise correlation of RDMs of each task with the $19$ other tasks to get a $20\times20$ task similarity matrix (Figure ~\ref{fig3}a). We perform a  hierarchical clustering from the similarity matrix, to visualize if the clustering groups the tasks according to the task type or some other criteria. We report the results in the experiments section and compare it with the clustering obtained with the Taskonomy approach.

We note that RSA is symmetric, as compared to the transfer performance based metric in Taskonomy~\cite{zamir2018taskonomy}. Yet, symmetry does not affect task similarity rankings, as the positions of the tasks in the rankings are computed by relative comparison, and therefore,  independent of symmetry.

\subsection{Does ranking using RSA depends on dataset and model size?}
We analyze whether RSA based task similarity depends on the model size and amount of training data. Intuitively, it should be independent of model and dataset size, because our method is based on relative similarities. To investigate this, we select a subset of Taskonomy tasks (details in supp. material section S1) and trained smaller models, one per task, with fewer parameters than the models provided by Taskonomy, and on a small subset of Taskonomy data.  First, we evaluate if we obtain a similar task clustering using the small models on the selected tasks. Then, for each small model, we compute the similarity score with the pre-trained Taskonomy models on all $20$ tasks. The same analysis is repeated with pre-trained Taskonomy model trained on the same task, and we compare the relative similarity based rankings of the small and Taskonomy high-performing models. If the relative rankings of both small and Taskonomy model are similar, then the result suggests that for a completely new task one can train a small model and compute similarity scores to rank them.

\subsection{Is RSA related to transfer performance?}
We investigate if RSA based task similarity can be applied to transfer learning problem. We first compute the correlation between each column of Taskonomy affinity matrix with RSA matrix after removing the diagonal. As the Taskonomy affinity matrix is populated by raw losses/evaluations, it is indicative of transfer learning performance~\cite{zamir2018taskonomy}. We next select a task and dataset different from Taskonomy and obtained the similarity scores of a model trained on the new task with Taskonomy pre-trained models. The pre-trained models were ranked according to the similarity score. We then use the pre-trained models for initializing the model and add the last task dependent layers on top of the initialized model to train on the new task. The ranking based on the transfer performance is compared with the ranking based on RSA to evaluate the relation between transfer performance and RSA. As we will see in the results, RSA can be used to select the high performing models for transfer learning.

%%%%%%%%%%%%%%%%%%%%%
% EXPERIMENTAL SET-UP
%%%%%%%%%%%%%%%%%%%%%
\section{Experimental set-up}

We first provide the details of datasets used for the experiments, followed by the details of the models' architecture. 

\subsection{Datasets}
\vspace{-2mm}
\paragraph{Taskonomy dataset} It includes over $4$ million indoor images from $500$ buildings with annotations available for $26$ image tasks. $21$ of these tasks are single image tasks, and $5$ tasks are multi-image tasks. For this work, we select $20$ single image task for obtaining task similarities$^1$. 

We randomly selected $500$ images from the Taskonomy training dataset as $500$ different conditions to perform RSA. These images are used as input to generate representations of different task-specific models to compute the RDMs.
%by computing pairwise dissimilarities between each pair of conditions.

To analyze the dependency of RSA on dataset and model size used for training, we select one building (Hanson) from Taskonomy dataset, which contains $12138$ images. We divide them into $10048$ training and $2090$ validation images. 

\vspace{-5mm}
\paragraph{Pascal VOC semantic segmentation} To evaluate the application of RSA in transfer learning, we select the Pascal VOC~\cite{Everingham10,hariharan2011semantic} dataset for semantic segmentation task. It has pixelwise annotations for $10,582$ training images,$1,449$ validation and $1,456$ test images. We argue that this task is different from the Taskonomy semantic segmentation as the images are from a different domain.

\subsection{Models}
Below, we provide details of the network architectures of pre-trained Taskonomy models, small models trained for Taskonomy tasks, and models used for Pascal VOC.

\paragraph{Taskonomy models}
The Taskonomy models~\footnote{publicly available at https://github.com/StanfordVL/taskonomy} consist of an encoder and decoder. The encoder for all the tasks is a Resnet-50~\cite{he2016deep} model followed by convolution layer that compresses the channel dimension of the encoder output from $2048$ to $8$. The decoder is task-specific and varies according to the task. For classification tasks and tasks where the output is low dimensional the decoder consists of 2-3 fully connected (FC) layers. For all the other tasks, the decoder consists of 15 layers (except colorization with 12 layers) consisting of convolution and deconvolution layers.

We select the final compressed output of the encoder as the representation for RSA as in~\cite{zamir2018taskonomy}. In Taskonomy approach, the compressed output of the encoder was used as an input to transfer function to evaluate the transfer learning performance.  Selecting the compressed output of the encoder ensures that the architecture for all the task is the same, and the differences in representation can only arise due to the task that the model was optimized for, as images are also the same for all tasks. 

We also explore the representation of earlier layers of the encoder and the task labels as the representation for computing RSA based similarity score. We perform this analysis to investigate how task specificity varies across the depth in the network and if the task's labels are enough to understand the relationship between tasks.
 
\vspace{-1mm}  
\paragraph{Small models}
The smaller version of the models follows a similar style to Taskonomy and consists of an encoder and decoder. The encoder consists of 4 convolution layer each with a stride of $2$ to generate a final feature map with the dimensions same as that of Taskonomy encoder. 
%The decoder is task-specific and varies depending on the task. For classification tasks and the tasks where the output is low dimensional the decoder consists of 2-3 fully connected (FC) layers. 
For this experiment, we select the tasks which require a fully-convolution decoder structure and use $4$ convolution layers each followed by an upsampling layer. The models are trained on Hanson subset of Taskonomy dataset. 
%The task-specific loss functions used for optimization and input/target preprocessing is the same as in Taskonomy implementation. 

\vspace{-5mm}
\paragraph{Pascal VOC Models}
We use two types of models for Pascal VOC semantic segmentation task: 1) a small model to compute similarity score with pre-trained Taskonomy models; 2) models initialized with pre-trained Taskonomy encoders to evaluate transfer learning performance. The small model consists of an encoder and a decoder. The encoder consists of 4 convolution layer each with a stride of 2 to generate a final feature map with the dimensions same as that of Taskonomy encoder. The decoder is an Atrous Spatial Pyramid Pooling (ASPP)~\cite{chen2018deeplab}, which contains convolution layers that operate in parallel with different dilations. The model is trained on Pascal VOC training set with learning rate $10\textsuperscript{-4}$ for $200,000$ iterations. The encoder representation of the small model trained on Pascal VOC is then used to compute similarity with Taskonomy pre-trained models. The models for evaluating transfer learning performance consists of an encoder with similar architecture as Taskonomy models and an ASPP decoder. The encoder part is initialized by the pre-trained Taskonomy models of the corresponding task.
%TODO: add plots with different iteration in the supplementary. Not mentioned here, but in the results section.

\vspace{-5mm}
\paragraph{Implementation and evaluation  details} We use the publicly available tensorflow implementation \footnote{https://github.com/sthalles/deeplab\textunderscore v3} of deeplabv3 \cite{chen2017rethinking} and modify the code for transfer learning experiments. We use RSA Matlab toolbox~\cite{nili2014toolbox} for RSA related analysis\footnote{Code available at https://github.com/kshitijd20/RSA-CVPR19-release}. We refer to the supplementary material for further details.  

\begin{figure}[t]
\begin{center}
    \includegraphics[width=1\linewidth]{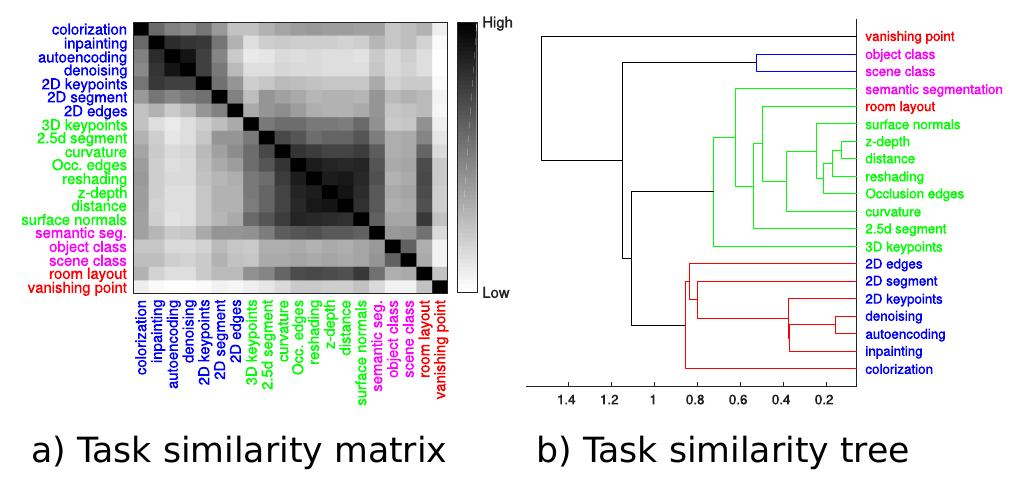}
\end{center}
   \caption{\textbf{Task similarity using RSA: a)} Similarity matrix of the $20$ Taskonomy tasks, \textbf{b)} Agglomerative clustering using RDM. \vspace*{-3mm}}
\label{fig4}

\end{figure}
\begin{figure*}
\begin{center}
    \includegraphics[width=1\linewidth]{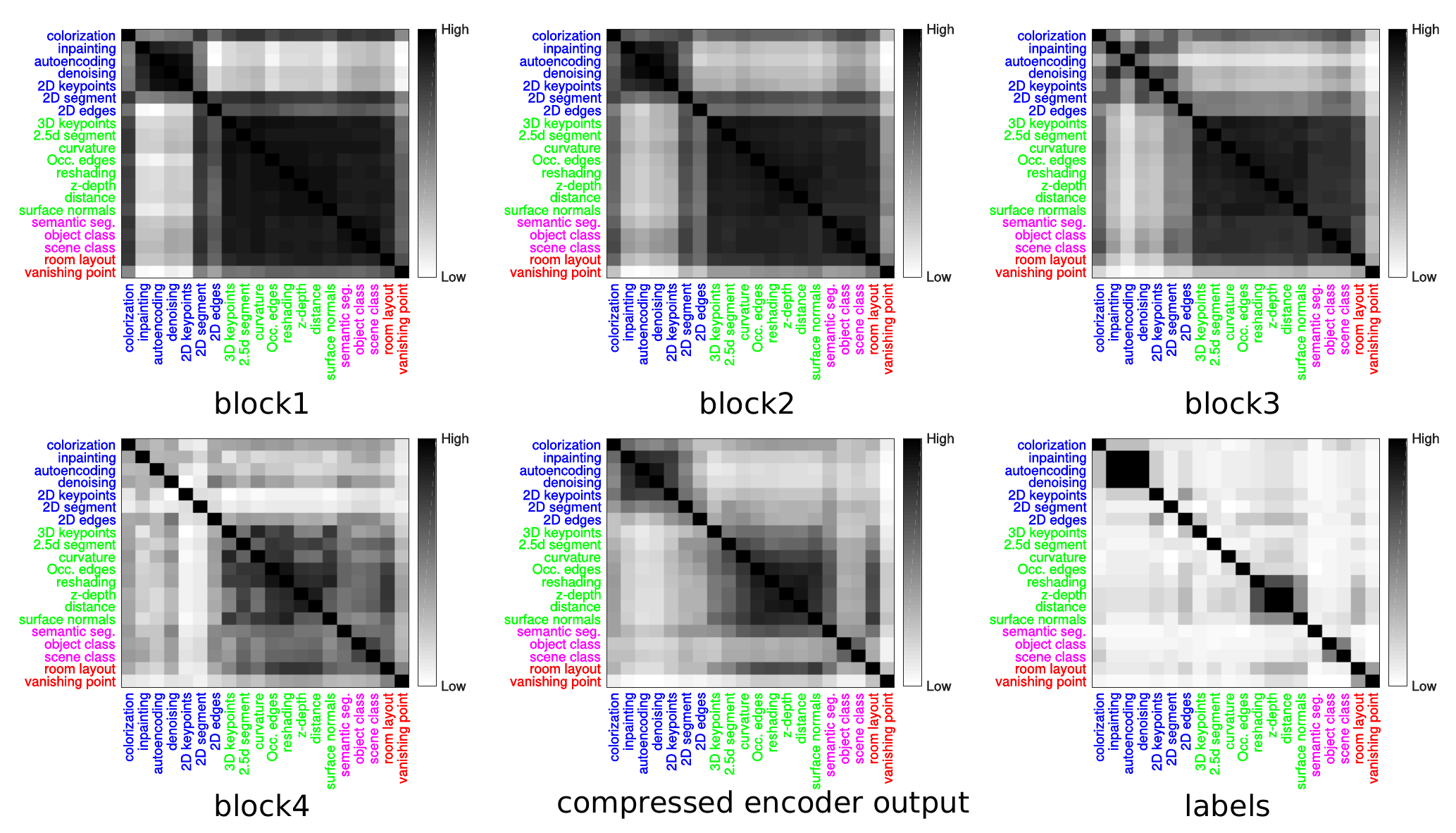}
\end{center}
   \caption{\textbf{Task taxonomy using RSA:  $1-5$)} Similarity matrix of $20$ Taskonomy tasks using features at different depth in the model as task-specific representations \textbf{$6$)} Similarity matrix of $20$ Taskonomy tasks using labels as task-specific representations.\vspace*{-2mm}}
\label{fig5}
\end{figure*}

\begin{figure}
\begin{center}
    \includegraphics[width=1\linewidth]{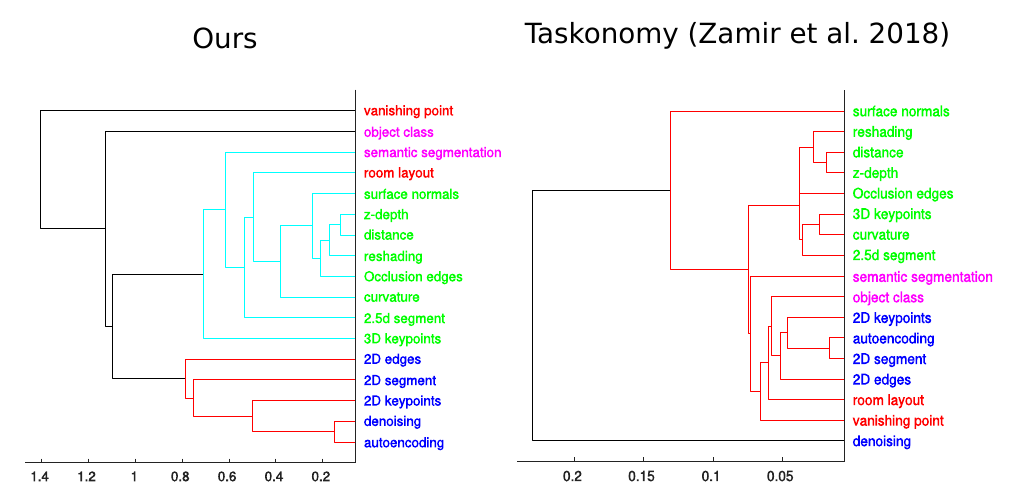}
\end{center}
   \caption{\textbf{RSA vs Taskonomy:} Clustering comparison. \vspace*{-4mm}}
\label{fig6}
\end{figure}

%%%%%%%%%%%
% RESULTS
%%%%%%%%%%%

\section{Results}
Here, we present the results of RSA for computing task similarity and its relation to transfer learning performance. We follow the same nomenclature of task type as in~\cite{zamir2018taskonomy}, and color code \textcolor{blue}{2D}, \textcolor{green}{3D}, \textcolor{magenta}{semantic}, and \textcolor{red}{geometric} tasks. 

\subsection{Task similarity using RSA}
Figure~\ref{fig4}a shows the similarity matrix of the tasks computed using RSA with the compressed encoder output as the task representation. Recall that we compute the $20\times20$ similarity matrix using RSA with given task-specific representations for all the randomly selected $500$ images. To visualize the relationship between tasks, we applied agglomerative hierarchical clustering%\footnote{We use default Matlab dendrogram function.}
to the similarity matrix. The resulting dendrogram from this clustering is shown in Figure~\ref{fig4}b. We can see that the tasks are clustered following visual criteria of 2D, 3D, and semantic tasks. 

We further investigate the task similarity using RSA at different depths in the encoder architecture and task labels. Figure~\ref{fig5} shows the task similarity matrix for different depths of the Resnet-50 encoder, namely blocks 1, 2, 3 and 4. We also compare the similarity matrix computed using the tasks' labels. 
We observe, in Figure~\ref{fig5}, that at block 1 all the similarity values are very high implying that at initial layers representations of most of the tasks are similar irrespective of the task type. As we go deeper, the similarity score between tasks starts decreasing, and in compressed encoder output, we can see three dark blocks corresponding to 2D, 3D, and semantic tasks. The above results further validate our choice of using compressed encoder output as the task-specific representation for assessing the similarity between tasks. Interestingly, the clustering using task labels does not group into tasks of the same type, and most of the similarity scores are low. Instead, the labels clustering follows the output structure of the labels, independently of the task type. This is because the labels contain only limited information about the task, and it depends on the annotator criteria on how to represent the output. 

 We next compare our approach with Taskonomy approach\footnote{We show 17 tasks as we had access to only affinity values of these tasks. For comparison with figure 13 in~\cite{zamir2018taskonomy}, please refer to section S2 of supplementary material}. We use hierarchical clustering to visually compare the dendrograms obtained using both the methods in Figure~\ref{fig6}.  For quantifying the similarity, we compute the correlation of Taskonomy similarity matrix with RSA similarity matrix ($\rho$ = 0.62, $r_{s}$ = 0.65). The results show that both approaches group the tasks into similar clusters with few exceptions.
 %TODO: add correlation values here
Room layout is grouped with the vanishing point in Taskonomy approach and in 3D tasks with our approach. Denoising is clustered with inpainting and autoencoding using our approach, which are related tasks. We argue that our results are plausible. %We argue that Taskonomy approach might have poor transfer learning performance for denoising as the input image should have noise, which is not included for transfer learning to other tasks, resulting in denoising not being clustered into any group. 

\subsection{Does model size impact similarity score? }
\begin{figure}[t]
\begin{center}
    \includegraphics[width=1\linewidth]{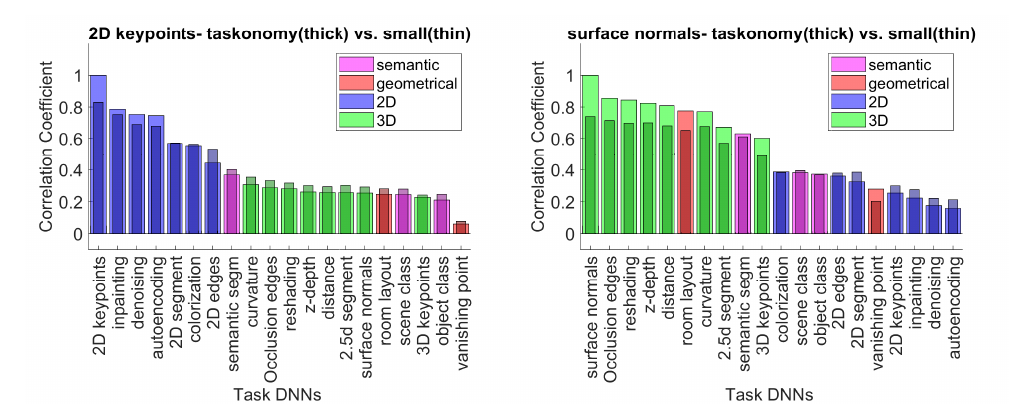}
\end{center}
   \caption{\textbf{Task taxonomy using small models:} Similarity ranking of \textbf{(a)} keypoint2d Taskonomy model vs small model. \textbf{(b)} surface normals Taskonomy model vs small model. \vspace*{-4mm}}
\label{fig7}
\end{figure}

In this experiment, we investigate how the model and dataset size affect task similarity. We show the results of similarity rankings for $2$ tasks: 2D keypoints and surface normals (for other tasks, please see section S1 in supplementary material). We compare the similarity rankings obtained using the small model trained on Hanson subset of Taskonomy data with the Taskonomy model trained on the same task. As we visually observe from the comparison (Figure~\ref{fig7}) in both the tasks the ranking look similar. For all the tasks considered in the above comparison the mean correlation is  high ($\rho$ = 0.84, $r_{s}$ = 0.85). 

Next, we also computed task similarity matrices by comparing a small model with small models trained on other tasks. We find that the correlation ($\rho$ = 0.85, $r_{s}$ = 0.88) between task similarity matrices (Figure S3) using Taskonomy model and small model is comparable to previous correlation results. %for 2D keypoints is $0.9873$, and for surface normals is $0.9818$. 
%TODO: fill the above and below correlation values
The above results together provide strong evidence that the model and dataset size do not have much effect on the similarity score.
%TODO: check consistency with results of new experiments. Do we need to mention/change something here?

\subsection{Model selection for transfer learning}
We first report the model selection using RSA for Taskonomy tasks and then on Pascal VOC semantic segmentation task.
\vspace*{-5mm}
\paragraph{Taskonomy}
 We obtain high mean correlation  ($\rho$ = 0.70, $r_{s}$ = 0.76) between RSA and transfer learning for 17 tasks from the Taskonomy dataset. We also report in Table~\ref{table1} that for 16 out of 17 tasks, the best model selected by RSA for transfer learning is in top-5 models selected using Taskonomy approach (transfer learning performance).

\begin{table}
\begin{center}
\begin{tabular}{|l|l|l|}
\hline
Top-1 & Top-3 & Top-5 \\
\hline
7/17 &   14/17 & 16/17   \\
\hline
\end{tabular}
\end{center}
\caption{Number of tasks for which best model selected for transfer learning using RSA is in top-n models according to transfer performance for 17 tasks\vspace*{-3mm}}
\label{table1}
\end{table}

\vspace*{-4mm}
\paragraph{Pascal VOC}
We show the relation of similarity score using RSA with transfer learning by selecting a new task (semantic segmentation in Pascal VOC). We compare the transfer learning performance of models initialized by different task-specific pre-trained models from Taskonomy dataset. Then we compare the transfer learning performance based ranking with similarity score ranking. Here we select the small Pascal model to compute the similarity with the Taskonomy models.  We report the robustness of similarity ranking using RSA with respect to model size, number of images used for RSA analysis, and different training stages in supplementary section S3. %We report the similarity ranking obtained using a large model similar to Taskonomy encoder architecture trained on the new task in S3 of the supplementary material and report the correlation of similarity ranking obtained using small and Taskonomy encoder ($\rho$ = $0.95$, $r_{s}$ = 0.96). We also compare the similarity rankings at different training stages (see S3 in supplementary materials) and observe that results show high correlation ($\rho>0.88$, $r_{s}>0.86$) across different iterations. The results suggest that we can use a small model with fewer iterations for computing task similarity.   

\begin{figure}[t]
\begin{center}
    \includegraphics[width=1\linewidth]{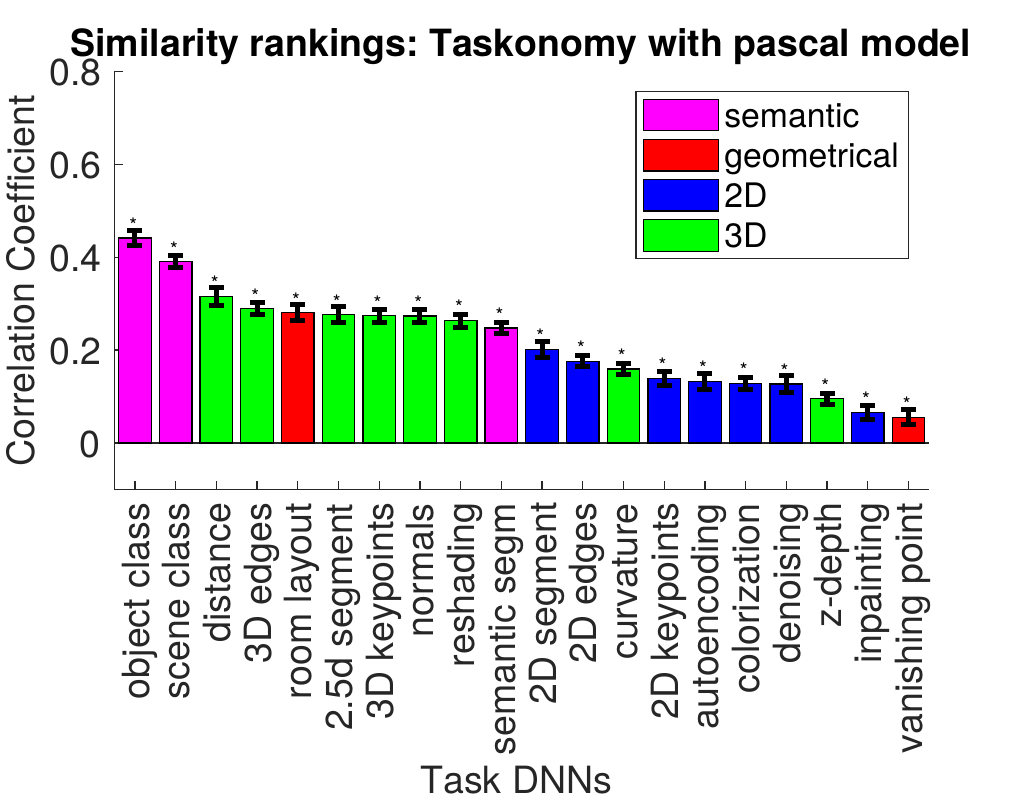}
\end{center}
   \caption{RSA based similarity of scores of pre-trained Taskonomy models with the small model trained on Pascal VOC. }
\label{fig8}
\end{figure}

%\paragraph{Semantic Segmentation on Pascal VOC}
We show the similarity score based ranking in Figure~\ref{fig8}. Surprisingly, semantic segmentation model from Taskonomy shows a lower similarity score as compared to other models trained on semantic (scene class, object class) and 3D tasks (occlusion edges, surface normals). Most of the 2D tasks show low similarity scores.

To investigate if similarity scores are related to transfer learning performance we evaluated the models initialized with task-specific Taskonomy models, finetuned with Pascal VOC training set, and compared the performance on Pascal VOC test set. Table~\ref{table1} shows the comparison of transfer learning performance for models with initialization from a set of selected tasks (For a complete comparison refer to section S3 in the supplementary material). The tasks are listed in the order of their similarity scores. We note from the table that the tasks on the top (object class, scene class, occlusion edges, and semantic segmentation ) shows higher performance while autoencoder and vanishing point performance is even less than model trained from scratch (random in Table~\ref{table2}). We note that our results are comparable to the results ($64.81\%$) reported in \cite{chen2017rethinking}, when they use Resnet-50 trained on Imagenet for initialization. %With modifications proposed in \cite{chen2017rethinking}, we argue that our results will be comparable to the numbers reported in \cite{chen2017rethinking}. 
The results provide evidence that the similarity score obtained using RSA provide an estimate of the expected transfer performance.

\begin{table}
\begin{center}
\begin{tabular}{|l|c|}
\hline
Initialization(Task) & mIoU \\
\hline
Object class &   0.6492   \\
Scene class & 0.6529 \\
Occlusion edges & 0.6496 \\
Semantic segmentation & 0.6487\\
Autoencoder & 0.5901 \\
Vanishing point & 0.5891 \\
\hline
Random(Taskonomy encoder) &  0.6083\\
Random(Small encoder) & 0.4072\\
\hline
\end{tabular}
\end{center}
\caption{Transfer learning performance on Pascal VOC test set.\vspace{-3mm}}
\label{table2}
\end{table}

%%%%%%%%%%%%%
% CONCLUSIONS
%%%%%%%%%%%%%

\section{Conclusion}
We presented an efficient alternative approach to obtain the similarity between computer vision models trained on different tasks using their learned representations. Our approach uses RSA, and it is suitable for obtaining task similarity by just using the pre-trained models without any further training, as opposed to the earlier state of the art method Taskonomy for this problem. 

We provided strong evidence that for obtaining the similarity, the model and training dataset size does not play a significant role and we can obtain a task similarity relative ranking using small models as well as state of the art models with few data samples. This comes with computational and memory savings. 

We also showed the relationship of the task similarity using RSA with the transfer learning performance and its applicability. We demonstrated on both, Taskonomy and Pascal VOC semantic segmentation, that the transfer learning performance is closely related to the similarity obtained with RSA. The above results showed that for domain shift the model trained on the same task might not be the best fit for transfer learning and our proposed approach can help in model selection for transfer learning. 
Our method is applicable to a wide range of potential problems, such as multitask models, architecture selection. 

%%%%%%
%END
%%%%%%
\vspace*{-3mm}
\paragraph{Acknowledgements}
This work was funded by the SUTD-MIT IDC grant (IDG31800103). K.D. was also funded by SUTD Presidents Graduate Fellowship. We thank Taskonomy authors for the support and the code.
{\small
\bibliographystyle{ieee_fullname}
\bibliography{egbib}
}
\include{supp-content}

\end{document}

%% file: supp-content.tex
\renewcommand{\thesection}{S\arabic{section}} 
\renewcommand{\thefigure}{S\arabic{figure}} 
\renewcommand{\thetable}{S\arabic{table}} 
\setcounter{section}{0}
\setcounter{figure}{0}
\setcounter{table}{0}

%%%%%%%%% TITLE
\title{Supplementary materials: \\Representation Similarity Analysis \\ for Efficient Task Taxonomy \& Transfer Learning}

\author{Kshitij Dwivedi \; \; \; \; \; \;  \; \; \; \; Gemma Roig\\
Singapore University of Technology and Design\\
%8 Somapah Rd, Singapore 487372\\
{\tt\small kshitij\_dwivedi@mymail.sutd.edu.sg, 
gemma\_roig@sutd.edu.sg}}

\maketitle

%\thispagestyle{empty}

%%%%%%%%% ABSTRACT

\begin{figure*}[t]
\begin{center}
    \includegraphics[width=1\linewidth]{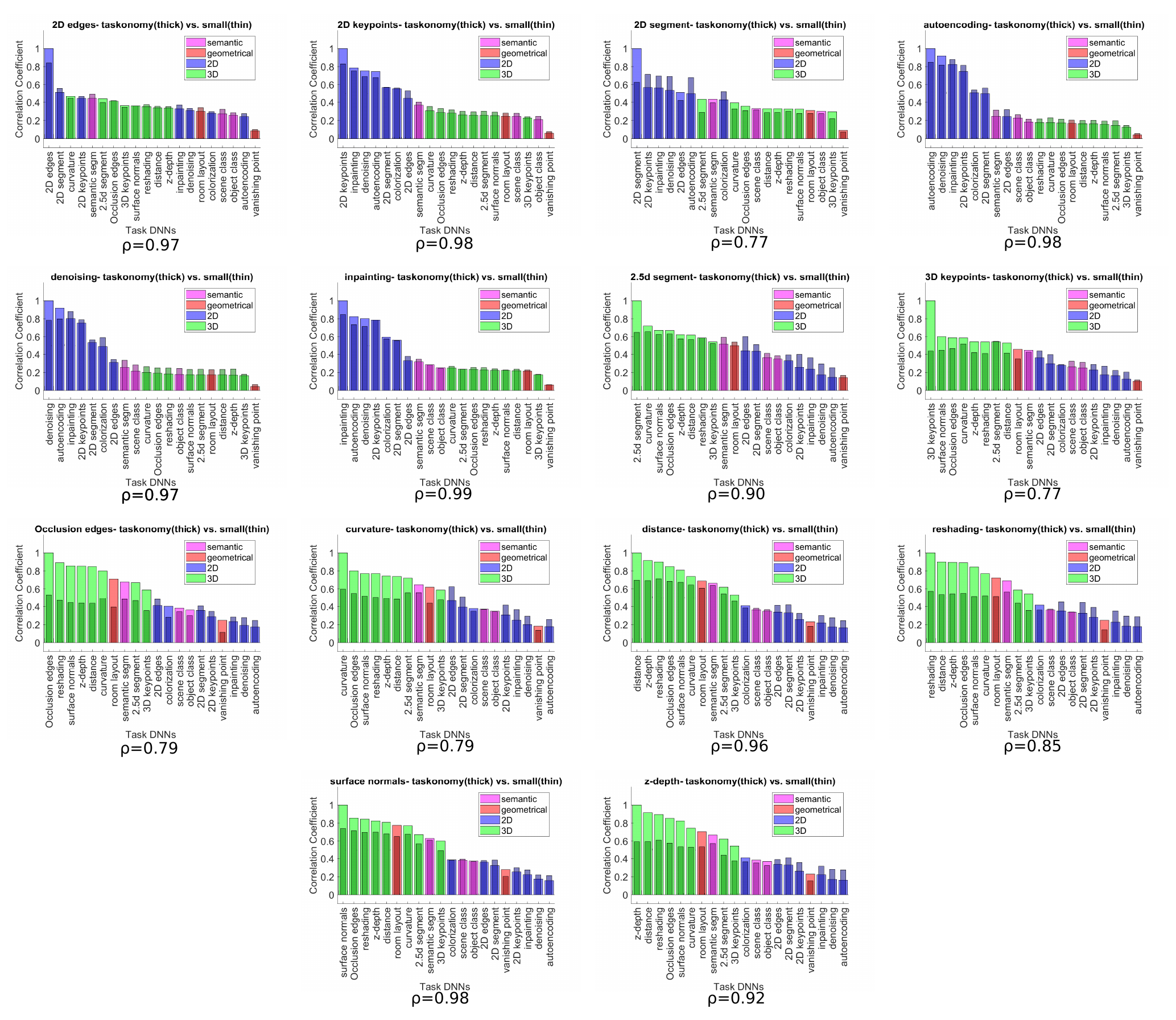}

\end{center}
\caption{\textbf{Similarity ranking with taskonomy model vs small models for 14 tasks.} The $\rho$ value below each plot specifies the Pearson's correlation coefficient between the two similarity rankings.  \vspace*{-3mm}}
\label{figs1}
\end{figure*}
\begin{figure*}[t]
\begin{center}
    \includegraphics[width=1\linewidth]{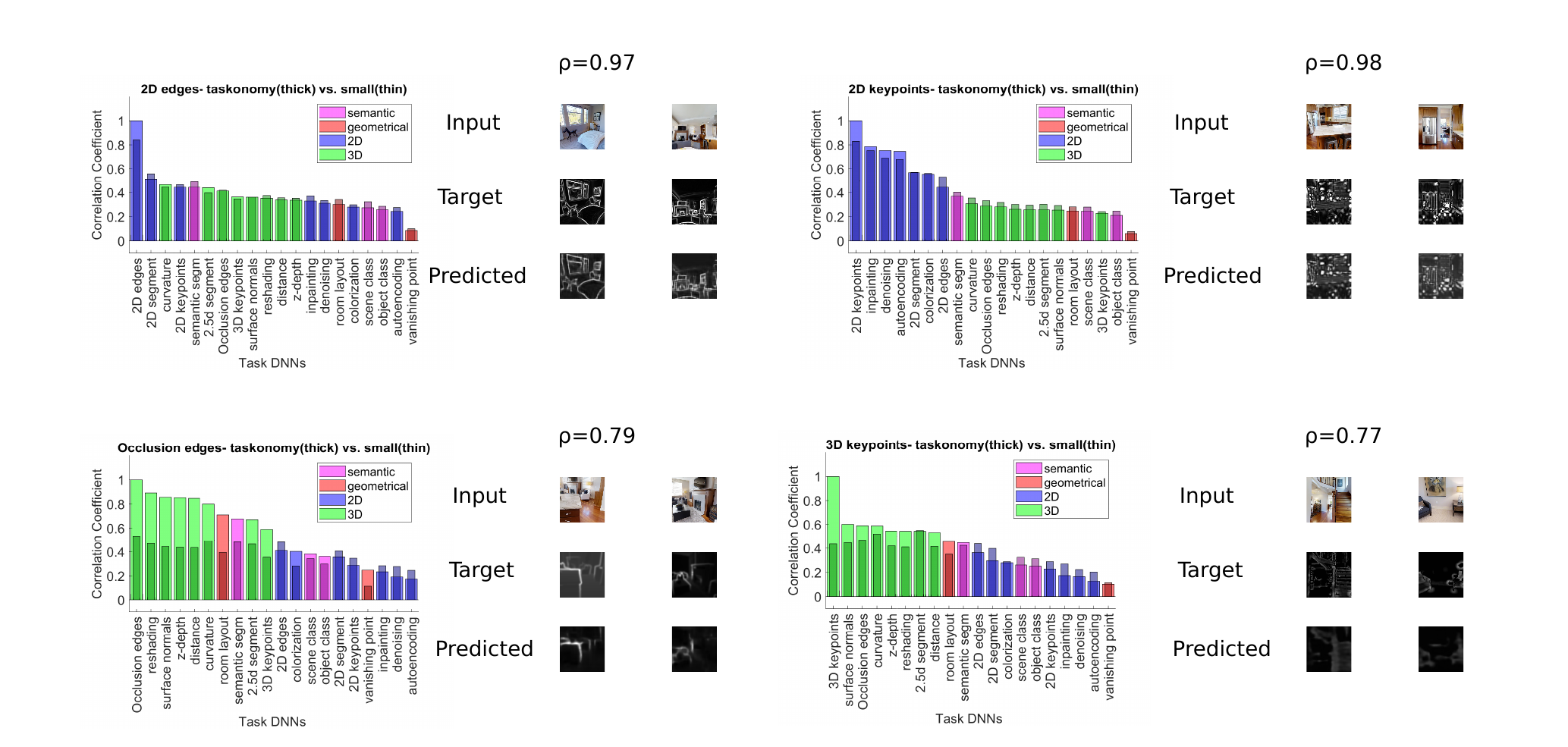}

\end{center}
\caption{\textbf{Is correlation related to visual similarity of the predicted output with the target?}  \vspace*{-3mm}}
\label{figs2}
\end{figure*}
%%%%%%%%% BODY TEXT

Here we report the additional details and results which we left in the main text to the supplementary material. In the first section, we provide details about the small models used and report the results and comparison with the Taskonomy pretrained models. In the second section, we compare the task similarity matrix and clustering using our RSA approach with that of Taskonomy\cite{zamir2018taskonomy} approach. In the third section, we report the consistency of RSA based similarity ranking and transfer learning performance for all the tasks. 
\section{Small models for task taxonomy}
We select the tasks (a total of $14$ tasks) which can be optimized using only L1/L2/triple-metric loss and the output of the task is spatial such that all the tasks can have the same decoder except the final layer. The architecture of the small model is reported in Table~\ref{tables1}. 

We show the task similarity comparison results (Figure~\ref{figs1}) of all the selected tasks. We note that for most of the 2D tasks the correlation (Pearson's $\rho$) of similarity rankings between small vs. Taskonomy models is very high ($\textgreater0.97$ except segment2d) and visually look similar. Although the correlation for all the 3D tasks is still high ($\textgreater0.77$), correlation values are relatively lower than 2D tasks.

We also evaluated the predicted output of 3D tasks and 2D tasks visually. We observed that for the tasks where the predicted output looks more similar to the target, the correlation is higher (Figure~\ref{figs2}). The difference in correlation could also be attributed to different training setting of Taskonomy and small models as it was not possible to exactly replicate the Taskonomy training with small models because the  training code is not publicly available, and the small models are trained using only a subset of the whole dataset.
\begin{table}
\begin{center}
\begin{tabular}{|l|c|c|c|}
\hline
Layer & Kernel size & \#
Channels & Stride \\
\hline
Encoder & & & \\
\hline
Conv1 &   $3\times3$ &16& 2  \\
Conv2 & $3\times3$ &32&2\\
Conv3 & $3\times3$ &64&2\\
Conv4 & $3\times3$&64&2\\
Conv5 & $3\times3$ &8&1\\
\hline
Decoder & && \\
\hline
Conv6 &$3\times3$&32&1\\
$Upscale\times2$ &&&\\
Conv7 &$3\times3$&16&1\\
$Upscale\times2$ &&&\\
Conv8 &$3\times3$&4&1\\
$Upscale\times2$ &&&\\
Conv9 &$3\times3$&4&1\\
$Upscale\times2$ &&&\\
Conv10 &$3\times3$&$n$&1\\

\hline
\end{tabular}
\end{center}
\caption{Small model architecture.The number of channel in Conv10 $n$ was task-specific\vspace{-3mm}}
\label{tables1}
\end{table}
\begin{figure}[t]
\begin{center}
    \includegraphics[width=1\linewidth]{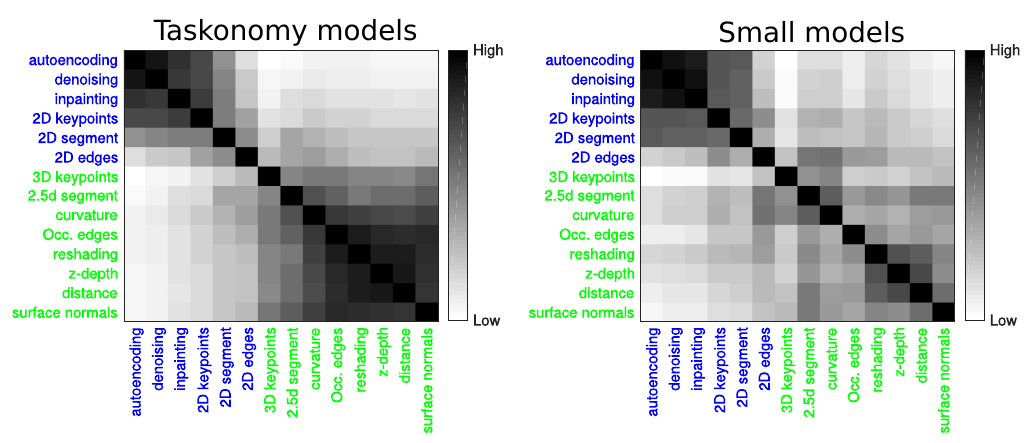}

\end{center}
\caption{\textbf{Task similarity matrix using Taskonomy models vs small models.}  }
\label{figs3}
\end{figure}
\begin{figure}[t]
\begin{center}
    \includegraphics[width=1\linewidth]{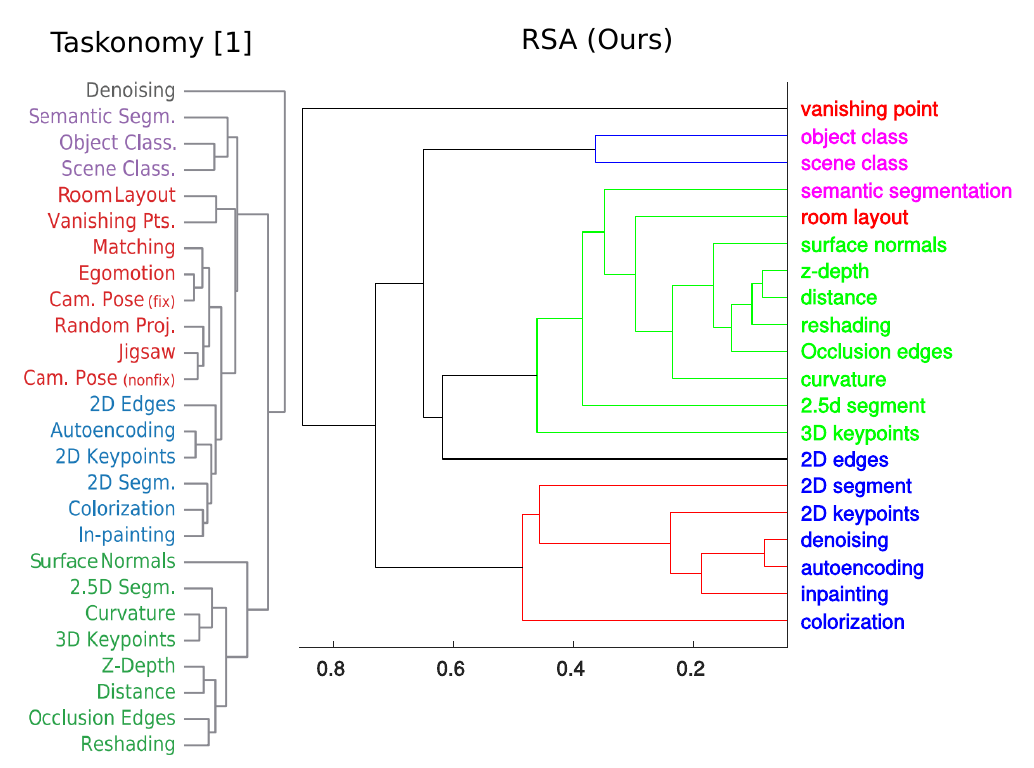}

\end{center}
\caption{\textbf{Clustering: Taskonomy vs RSA (Ours)} Image source: Figure 13 from ~\cite{zamir2018taskonomy} }
\label{figs4}
\end{figure}
\begin{figure}[t!]
\begin{center}
    \includegraphics[width=1\linewidth]{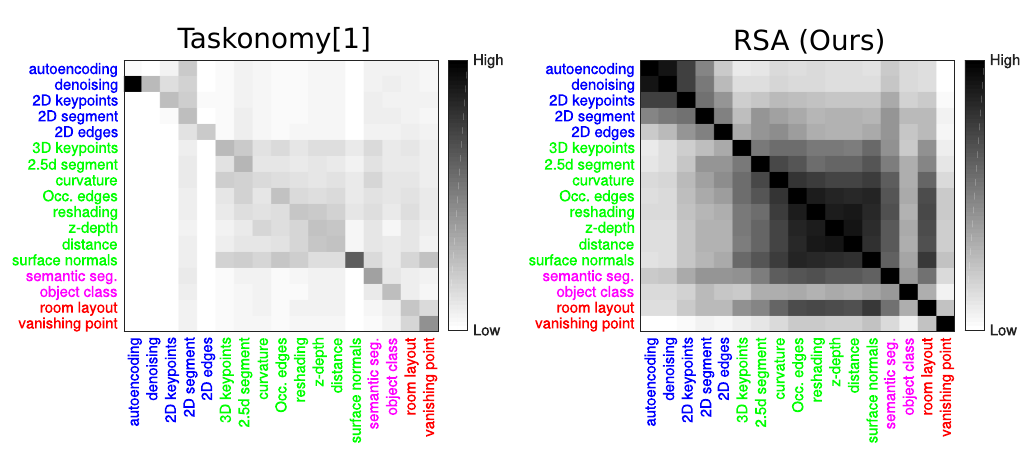}

\end{center}
\caption{\textbf{Similarity matrix: Taskonomy vs RSA(Ours)}  }
\label{figs5}
\end{figure}
We computed the task similarity matrix for the selected tasks using both small models and Taskonomy models. Although the similarity ranking using small models on 3D task did not show as high correlation with the Taskonomy models,  we found that the Pearson's correlation between them is high (0.8510). On visual inspection of both similarity matrices (Figure~\ref{figs3}), 2D tasks of small models show similar scores as with Taskonomy models. The 3D tasks although show higher similarity with corresponding 3D tasks rather than 2D tasks but similarity scores within 3D tasks are lower and therefore matrix looks lighter as compared to the similarity matrix with Taskonomy models.

\section{Taskonomy\cite{zamir2018taskonomy} vs RSA(Our approach)}

We show the clustering obtained using Taskonomy approach and compare it our approach in Figure~\ref{figs4}. From the figure, we observe that almost all of the 20 single image task we select for our paper (except room layout and denoise) belong in the same cluster as using Taskonomy approach. It is also possible that the difference in clustering arises due to different clustering method, which was not specified, used in ~\cite{zamir2018taskonomy}.

One other advantage of our approach over Taskonomy is that our similarity scores lie between -1 and 1 and thus similarity matrix is easy to visualize and evaluate. In Taskonomy approach, an exponential scaling of the similarity score has to be performed to bring them in a good range for visualization. Figure~\ref{figs5} shows both the similarity matrix without any scaling.

\section{Transfer learning in Pascal VOC}
In the first three subsections below, we show the consistency of RSA with varying number of iterations, the model size, and the number of images selected for RDM computation. In the last subsection, we report the transfer learning performance of all the task DNNs used for initialization. 
\begin{figure}[t]
\begin{center}
    \includegraphics[width=1\linewidth]{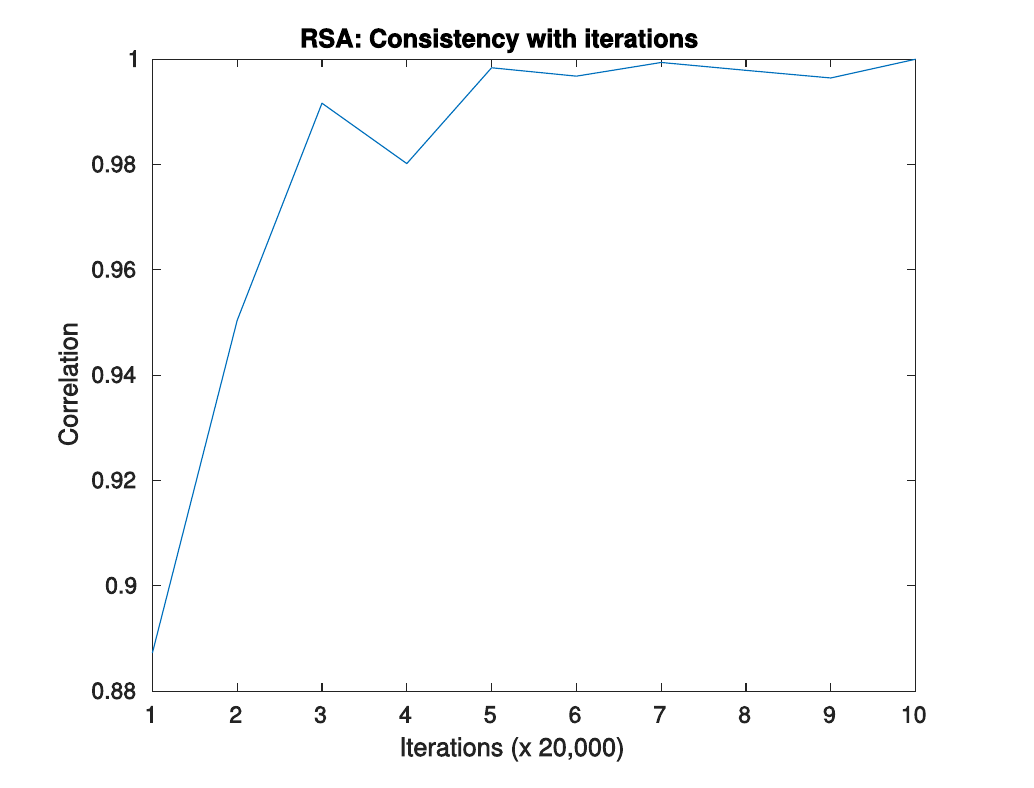}

\end{center}
\caption{\textbf{Consistency with training iterations}  }
\label{figs6}
\end{figure}
\subsection{Consistency with training stage}
We show in Figure~\ref{figs6} that even at $1/10$ of the final training stage the Pearson's correlation with the final stage is 0.88 and after $1/2$ of the training the correlation with the final stage stays above 0.99. This shows that one can also use models from an early stage of training for task similarity using RSA.
\begin{figure}[t]
\begin{center}
    \includegraphics[width=1\linewidth]{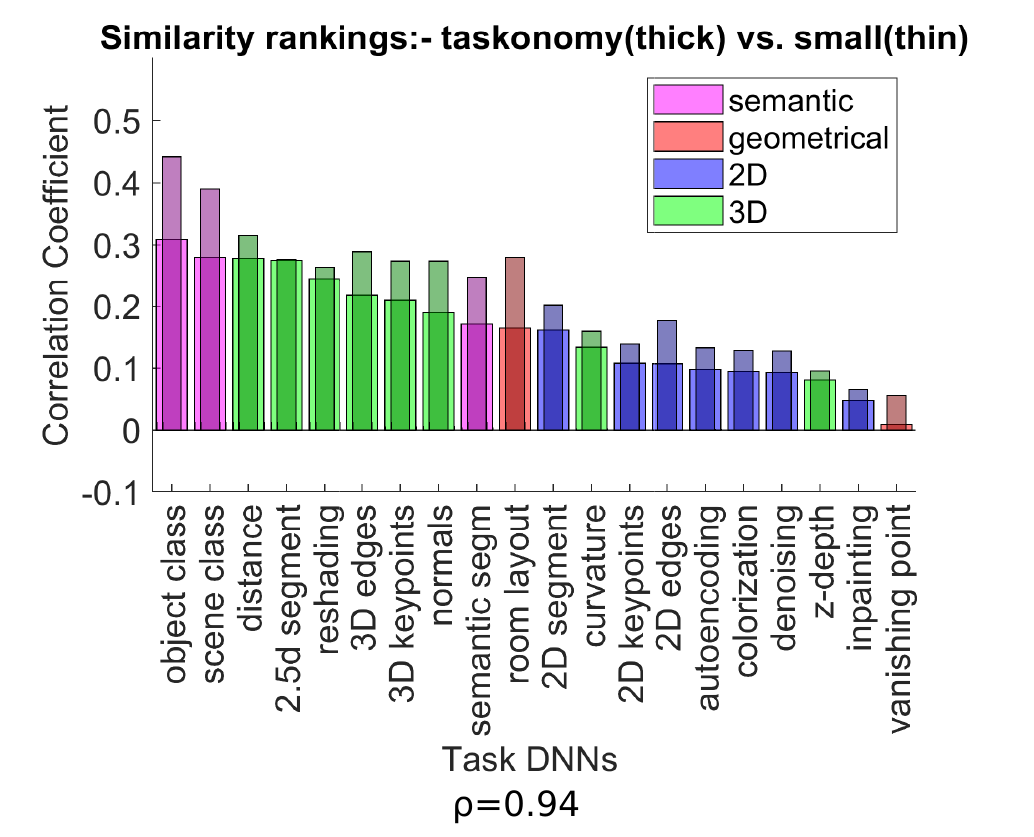}

\end{center}
\caption{\textbf{Consistency with model size}  }
\label{figs7}
\end{figure}
\begin{figure}[t]
\begin{center}
    \includegraphics[width=1\linewidth]{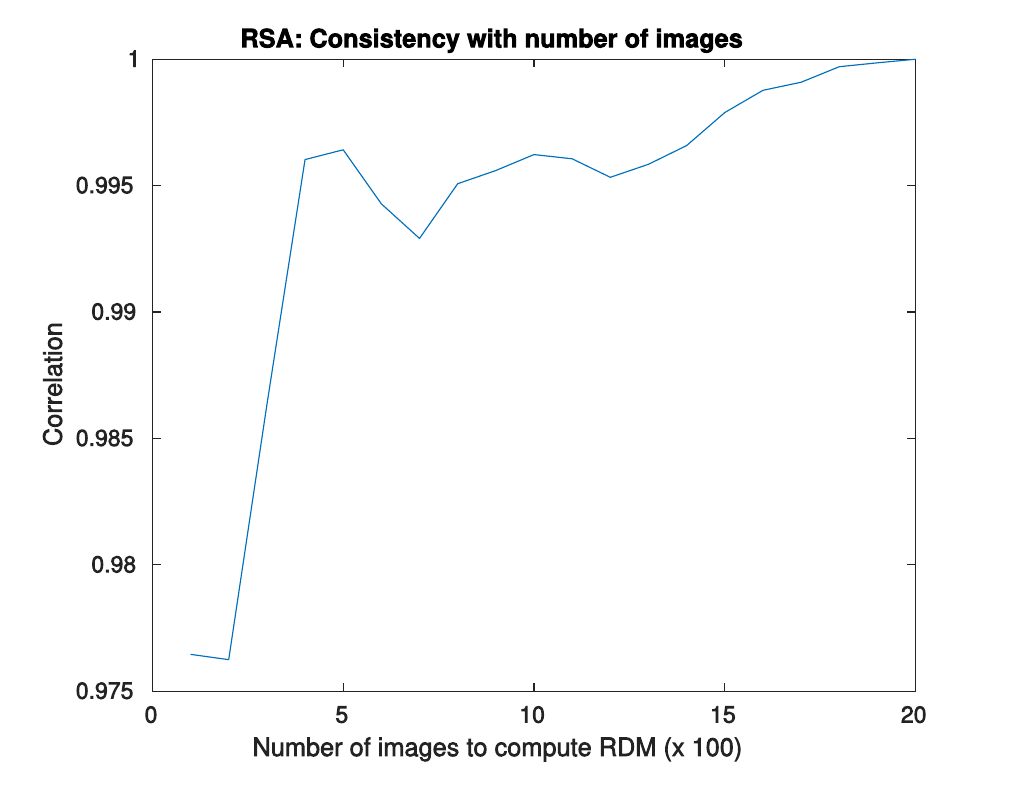}

\end{center}
\caption{\textbf{Consistency with number of images}  }
\label{figs8}
\end{figure}

\begin{figure}[t]
\begin{center}
    \includegraphics[width=1\linewidth]{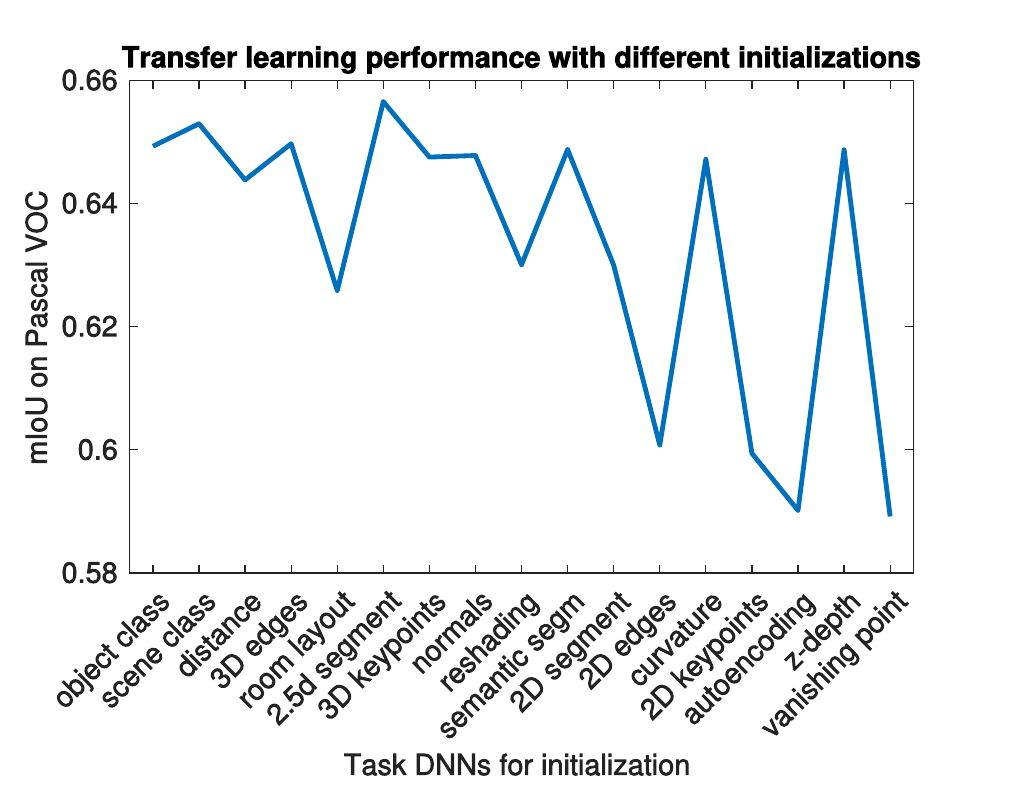}

\end{center}
\caption{\textbf{Transfer learning performance in descending order of similarity scores with task DNNs on the x-axis as initialization}  }
\label{figs9}
\end{figure}
\subsection{Consistency with model size}
We show in Figure~\ref{figs7} the comparison of task similarity obtained using a small encoder (thin bars) vs. task similarity obtained using taskonomy encoder architecture (thick bars). A high correlation ($\rho$ = $0.95$, $r_{s}$ = 0.96) suggests that we can use small models to train on a new task and use RSA to select a good model for initialization.

\subsection{Consistency with the number of images}
We varied the number of images from $100$ to $2000$ and plot the Pearson's correlation of task similarity ranking obtained using n images with the task similarity ranking obtained using $2000$ images (Figure~\ref{figs8}). After $400$ images the Pearson's correlation with the task similarity ranking is always above $0.99$, thus suggesting that around $500$ images are sufficient for RDM computation.

\subsection{Transfer learning performance for all the tasks}
Figure~\ref{figs9} shows the transfer learning performance (mIoU) for 17 single image tasks \footnote{ We ignore denoise, autoencoding, and colorization as these tasks require modified input} in the descending order of similarity rankings. The curve shows that the performance in most of the tasks seems to decrease as the similarity score decreases (although it is not a perfect monotonically decreasing curve).Also, generally the tasks with higher similarity ranking (object class, surface normals, segment25d) showed high transfer learning performance, and tasks with lower similarity score (autoencoding, vanishing point) showed lower performance.